\documentclass[runningheads]{llncs}

\usepackage{times}
\usepackage{soul}
\usepackage{url}
\usepackage[hidelinks]{hyperref}
\usepackage[utf8]{inputenc}
\usepackage[small]{caption}
\usepackage{graphicx}
\usepackage{amsmath}
\usepackage{booktabs}
\usepackage{algorithm}
\usepackage{algorithmic}
\usepackage{caption}
\usepackage{makecell} 

\usepackage{xcolor,colortbl}

\newcommand{\hide}[1]{}
\newcommand{\myModel}{\textsc{SimSkip}}

\begin{document}
\title{Can Contrastive Learning Refine Embeddings}


\author{Lihui Liu\thanks{Work conducted while the author was an intern at Amazon.}\inst{1}\orcidID{0000-0002-3758-4041} \and
Jinha Kim\thanks{Corresponding author}\inst{2}\orcidID{0009-0001-2758-7425} \and
Vidit Bansal\inst{2}\orcidID{0009-0008-4398-8617}}

\institute{University of Illinois at Urbana-Champaign, Champaign IL 61820, USA \email{lihuil2@illinois.edu} \and
Amazon, Seattle WA 98109, USA
\email{\{jinhak,bansalv\}@amazon.com}}

\maketitle
\begin{abstract}
Recent advancements in contrastive learning have revolutionized self-supervised representation learning and achieved state-of-the-art performance on benchmark tasks. While most existing methods focus on applying contrastive learning on input data modalities like images, natural language sentences, or networks, they overlook the potential of utilizing output from previously trained encoders. In this paper, we introduce \myModel{}, a novel contrastive learning framework that specifically refines the input embeddings for downstream tasks. Unlike traditional unsupervised learning approaches, \myModel{} takes advantage of the output embedding of encoder models as its input. Through theoretical analysis, we provide evidence that applying \myModel{} does not lead to larger upper bounds on downstream task errors than that of the original embedding which is \myModel{}'s input. Experiment results on various open datasets demonstrate that the embedding by \myModel{} improves the performance on downstream tasks.
\end{abstract}

\section{Introduction}

Embedding symbolic data such as text, graphs, and multi-relational data has become a key approach in machine learning and AI ~\cite{nickel2017poincare}. The learned embeddings can be utilized in various applications. For instance, in NLP, word embeddings generated by WORD2VEC~\cite{word2vec} or BERT~\cite{bert} have been employed in tasks like question answering and machine translation. In the field of graph learning, embeddings of graphs like NODE2VEC~\cite{node2vec} and DEEPWALK ~\cite{deepwalk} have been used for node classification and link prediction in social networks. Similarly, in computer vision, image embeddings such as ResNet~\cite{resnet} can be used for image classification.

Despite the progress in representation learning, learning effective embeddings remains a challenging problem. Especially for deep learning models, they often require a large amount of labeled training data, which can be costly and limit their applicability. These are known as supervised learning tasks. Additionally, the learned embeddings often perform well on one task but not on others. To address this issue, contrastive learning has been proposed as a solution that can learn good representations with only a limited amount of labeled data.

Contrastive learning has the advantage of being able to learn representations without label information, thus saving a significant amount of human effort and resources that would have been used for data labeling. The fundamental idea of contrastive learning is to bring together an anchor and a "positive" sample in the embedding space while pushing apart the anchor from many "negative" samples~\cite{simCLR}. As there are no labels available, a positive pair often consists of data augmentations of the sample, and negative pairs are formed by the anchor and randomly chosen samples from the minibatch ~\cite{simCLR}. Although the concept is simple, recent research has shown that contrastive learning methods can achieve comparable results to supervised methods ~\cite{NEURIPS2020_d89a66c7,robinson2021contrastive}.

Given the success of contrastive learning, a logical question to ask is whether using the output of another embedding model as input to contrastive learning can further refine the embedding space and make it perform better for downstream tasks. This is the question we aim to answer in this paper. We propose a new approach, called \myModel, that takes the output embedding of another model as input and applies contrastive learning on it. Our proposed method aims to fine-tune the input embedding space, making it more robust for downstream tasks. 
We theoretically prove that after applying \myModel\ on the input embedding space, for a downstream task, the error upper bound of the new learned fine-tuned embedding will not be larger than that of the original embedding space.
We conduct extensive experiments on various datasets and downstream tasks to evaluate the performance of our proposed approach and compare it with other state-of-the-art methods. The results show that the proposed \myModel\ can refine the input embedding space and achieve better performance on downstream tasks.

In summary, the main contributions of this paper are:
\begin{itemize}
  \item {\bf Problem Definition.} To the best of our knowledge, we are the first to propose and investigate the use of contrastive learning to improve the robustness of embedding spaces.
  \item {\bf Algorithm} We propose a skip-connection-based contrastive learning model, \myModel, and theoretically prove that it can reduce the error upper bound of downstream tasks.
  \item {\bf Empirical Evaluations}. We conduct extensive experiments on several real-world datasets and various downstream tasks. The results of our experiments demonstrate the effectiveness of the proposed \myModel.
\end{itemize}

\section{Preliminaries and Problem Definition}\label{problem-definition}



\subsection{Contrastive Learning}

Contrastive learning aims to learn effective representations by pulling semantically similar samples together and pushing dissimilar samples apart ~\cite{gao2021simcse}.
In self-supervised setting as such contrastive learning, constructing positive and negative pairs from unlabeld dataset through data augmentation is critical.
For example, in visual representations, an effective approach is to generate two augmented images from one input image and use them as the positive pair, while other images in the same mini-batch are treated as negative pairs of the input image. There are several different data augmentation methods such as cropping, flipping, distortion, and rotation ~\cite{simCLR}.
In node representations in graphs, one idea is to use the neighborhood of the given node as positive pairs, while nodes that are farther away are treated as negative pairs. For graph-level representations, operations such as node deletion and edge deletion can be used to generate positive augmentations of the input graph.

After building positive and negative pairs, neural network-based encoders are used to learn representation vectors from augmented data examples. Various network architectures such as ResNet ~\cite{simCLR} for images and BERT ~\cite{gao2021simcse} for text can be used. The output representation vectors of the encoders are used as the final embedding of the input data.
To learn an effective embedding space discriminating positive and negative pairs, a simpler neural network called projector is stacked on top of a encoder and the contrastive loss is applied against the projector output. A commonly used projector is an MLP with one or two layers, which is simple to implement.

In training, first, a random sample of $N$ examples is taken for a mini-batch.
Then, $N$ pairs are constructed from $N$ samples through data augmentation, which lead $2N$ examples total in the mini-batch.
$N$ augmented pair of an input data point are treated as the positive pair in the mini-batch.
For each augmented positive pair $(i, j)$, the remaining $2N - 2$ example are used to construct negative examples $(i, k)$. The commonly used contrastive learning loss is

\begin{equation}
l_{i,j} = - \textrm{log} \frac{\textrm{exp}(\textrm{sim}(z_i, z_j) / \tau)}{ \sum_{k=1}^{2N} \mathbf{I}_{  k \neq i,j } \textrm{exp}(\textrm{sim}(z_i, z_k) / \tau) }
\end{equation}
where $\tau$ is the temperature, $\textrm{sim}$ is a similarity function such as the cosine similarity, $z_i (= p(f(x_i))$ is the output of the projector $p$ which takes the output of the encoder $f$ ~\cite{simCLR}.

\subsection{Problem Statement}

In this paper, we focus on investigating whether contrastive learning can refine the embeddings for downstream tasks. Given an input dataset $D = \{d_i\}_{1}^{N}$, an arbitrary embedding function $h()$, and its output embedding  $\mathcal{X}$, where $x_i \in \mathcal{X}$ is the embedding of data point $d_i$ ($x_i = h(d_i)$), our goal is to design a new embedding function $f$ such that $f(h(d_i))$ performs no worse than $h(d_i)$ given an arbitrary downstream task $\mathrm{T}$.

\section{Method}\label{method}
In the previous sections, we outlined the concept of unsupervised contrastive learning. In this section, we will delve into the specifics of using \myModel\ that refines pre-existing embeddings.

\subsection{Contrastive Learning Limitation}

\begin{figure}
    \centering
    \includegraphics[width=0.4\textwidth]{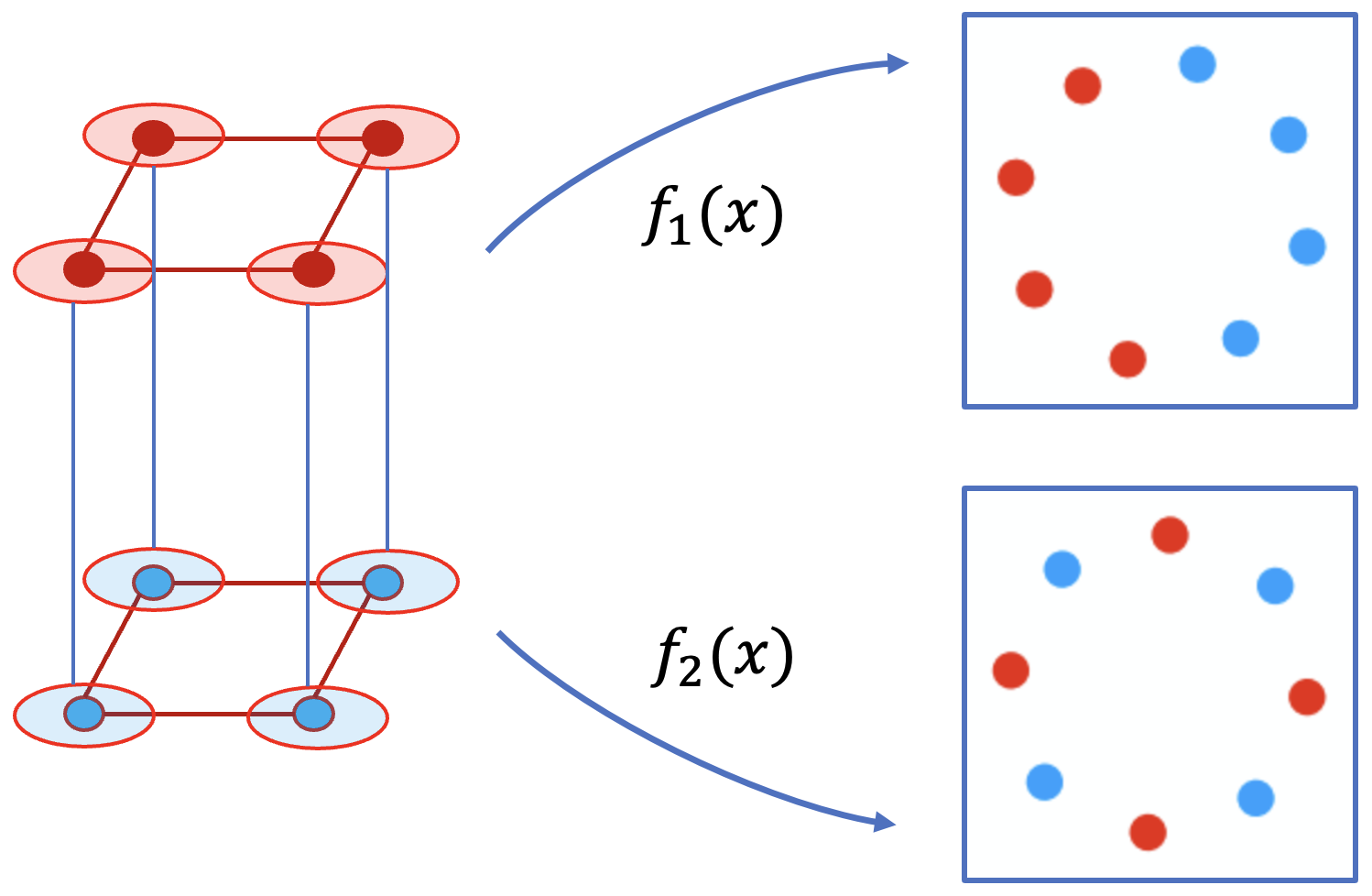}
    \caption{The problem of existing unsupervised contrastive learning}. 
    \label{problem_of_existing_method}
\end{figure}

The architecture of contrastive learning ensures that augmentations of the same data point are close to each other in the embedding space. However, this alone does not guarantee that the learned embeddings are suitable for downstream tasks. 
As shown in Figure ~\ref{problem_of_existing_method}, assuming we have 8 input embedding points that belong to two different classes, red and blue. When adding Gaussian noise to the original embedding points to create their augmentations, the augmented positive points are represented by the circles around the points on the left side of Figure ~\ref{problem_of_existing_method}.
When there are two different contrastive learning encoders, $f_1$ and $f_2$, they will map all augmentations of the same data point to the embedding points close to that of the original data point in the contrastive embedding space.
If the two augmentations are denoted as $x_{i1}$ and $x_{i2}$, $\textrm{sim}(f_1(x_{i1}), f_1(x_{i2}))$ will be close to 1 and the same for $f_2$. 
Since all the other augmentation examples are treated as negative examples, it is clear that the contrastive loss of $f_1(x)$ will be very similar as that of $f_2(x)$ when they map the original data points as shown in the right side of Figure~\ref{problem_of_existing_method}.

Even though the contrastive loss of $f_1$ and $f_2$ are very similar, the performance of the downstream classification task may differ between the two embedding spaces.
For example, $f_1$ separates the red and blue points into distinct clusters, which makes it easy for the downstream classification task to accurately classify them. However, in the embedding space created by $f_2$, the red and blue points are mixed together, which results in poor performance for the downstream classification task.

\subsection{\myModel\ Details}

\begin{figure*}
    \centering
    \includegraphics[width=1\textwidth]{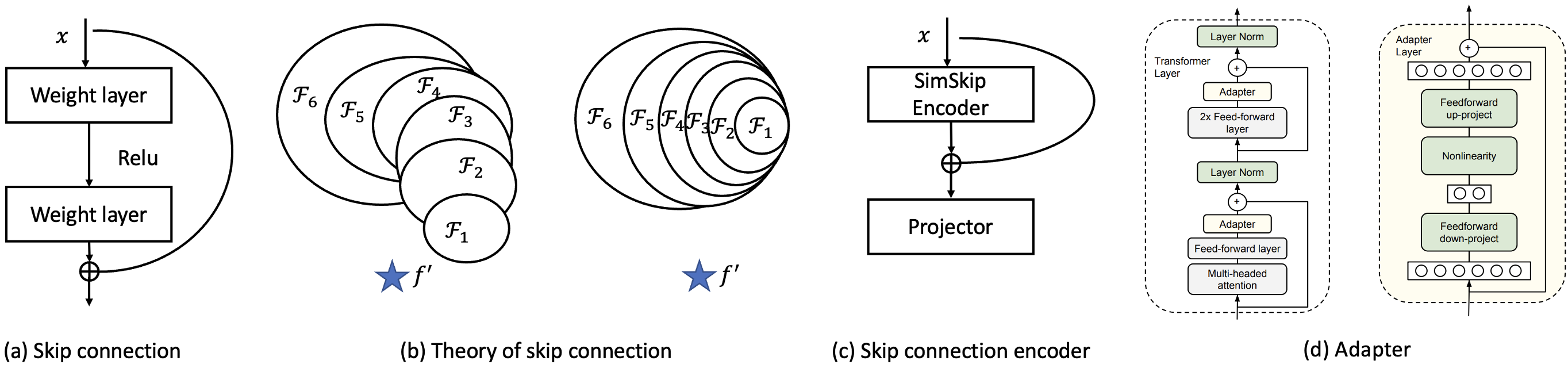}
    \vspace{-1\baselineskip}
    \caption{The Skip Connection. The picture of Adapter is from ~\cite{houlsby2019parameterefficient}.}
    \label{skip_connection}
    \vspace{-1\baselineskip}
\end{figure*}

To address this problem, we introduce skip connection contrastive learning. This method is similar in principle to ResNet ~\cite{resnet} as illustrated in Figure ~\ref{skip_connection}. The idea behind skip connections is that it retains the expressiveness of the original network. A specific network architecture defines a class of functions $F$ that it can represent.
Suppose $f^{'}$ is the optimal function we aim to find. If it is within $\mathcal{F}$, we are in good shape. However, it is often the case that it is not. Therefore, our goal is to find the best approximation of $f^{'}$ within $\mathcal{F}$.

A naive way to achieve this is to increase the depth and width of the neural network. By adding more layers and neurons, the network can represent a new class of functions $\mathcal{F}'$, which is more expressive than $\mathcal{F}$.
In general, we expect that $f_{\mathcal{F}'}$ would be better than $f_\mathcal{F}$, as a more expressive function class should be able to capture more complex patterns in the data.
However,
this may not be the case.
In fact, increasing the depth and width of the network can lead to
a worse $f_{\mathcal{F}'}$, as illustrated by Figure ~\ref{skip_connection} (b).
In this example, even though $\mathcal{F}_6$ is larger than $\mathcal{F}_3$, its optimal approximation is farther from the optimal function $f'$.

To solve this problem, Kaiming et al in ~\cite{resnet} proposes to use skip-connection to avoid the aforementioned issue from the non-nested function classes.
The idea of skip-connection is that it can create nested function classes where $\mathcal{F}_1 \subseteq ... \subseteq \mathcal{F}_6$ as shown on the right of Figure ~\ref{skip_connection} (b).
Because the larger function classes contain the smaller ones, it can guarantee that increasing them strictly increases the expressive power of the network.
For deep neural networks,
if we can train the newly-added layer into an identity function $f(x) = x$, the new model will be as effective as the original model.
As the new model may get a better solution to fit the training dataset, the added layer might make it easier to reduce training errors.

Building on the idea of incorporating skip connections, we propose a model named \myModel\ that utilizes contrastive learning to refine embedding based on the original input embedding. The architecture of \myModel\ is illustrated in Figure ~\ref{skip_connection} (c).
\myModel\ consists of two components: a skip connection based encoder and a projector. 
The detail of the encoder can be found in Figure ~\ref{architecture}.

The projector is a Multi-layer Perceptron (MLP) with one hidden layer, represented as $W_2 \sigma (W_1 x)$, where $\sigma$ is a ReLU non-linearity.
By incorporating skip connections, the expressive power of the network (contrastive learning encoder) is increased. Therefore, the new learned embedding should perform at least as well as the original embedding in downstream tasks.

\subsection{Data Augmentation}\label{data_aug}
Data augmentation is commonly used in contrastive learning to generate positive samples for a given data point. However, when the input to the model is the output embedding of another model, traditional data augmentation methods are not applicable. Image-based techniques like cropping, resizing, cut-out, and color distortion, as well as Sobel filtering, can only be applied to images ~\cite{simCLR}. Other methods such as node deletion and edge deletion for graphs are also not suitable for this purpose. How to design an effective data augmentation strategy is also crucial. 

Inspired by ~\cite{xxxxxfasdfasfs}, in this paper, we use two types of data augmentation to embedding output of an encoder network -- masking and Gaussian noise. 

\noindent{\bf A - Random Masking.}  Random masking is applied to the input embedding. Specifically, given an input embedding $\mathbf{e}_i \in R^{d}$, a random vector $M \in \{0, 1\}^d$ is created where 0 indicates that the element will be masked and 1 indicates no change. The number of 0s in $M$ is drawn from a Bernoulli distribution $B(1 - \alpha)$, where $\alpha$ is a hyperparameter. The output after applying random masking is $\mathbf{e}_i \circ M$, where $\circ$ represents element-wise multiplication.

\noindent{\bf B - Gaussian Noise.} When adding Gaussian noise to the input embedding $\mathbf{e}_i \in R^{d}$, a random vector $\epsilon \sim N(0, \textbf{I} )$ is first sampled from a multi-variable Gaussian distribution, where $\epsilon \in R^{d}$ and each element in $\epsilon$ is drawn from a Gaussian distribution with 0 mean and unit variance. The output after adding the Gaussian noise is $\mathbf{e}_i + \delta \circ \epsilon$, where $\delta$ is a hyper-parameter.

\subsection{Theoretical proof}

In this section, we theoretically demonstrate why \myModel{} may refine it's input embedding. Here, `refine' means that the embedding which \myModel{} produces has no worse downstream performance than that of the original embedding which is \myModel{}'s input.
We initially establish the upper bound for the loss in any downstream task within the context of contrastive unsupervised learning, as demonstrated in ~\cite{bound}. 
Then, we prove that using a skip connection-based network as the contrastive learning encoder can achieve a smaller or equal loss upper bound for downstream classification tasks compared to using original input embedding directly. 
\\

\noindent{\bf A - Preliminary.}
Let $\mathcal{X}$ denote the set of all possible data points.
Let $f_1(x)$ represent an arbitrary neural network that takes $x$ as its input.
Then, a neural network with skip connection $f_2$ can be denoted as 
\begin{align}\label{res_formula}
f_2(x) = f_1(x) + x = f_1(x) + f_I
\end{align}
where $f_I = x$.


\noindent{\bf B - Downstream Task Loss for Contrastive Unsupervised Learning.}
In this section, we present an upper bound for the loss of a supervised downstream task which uses representation learned by any contrastive learning, as originally shown in ~\cite{bound}.

In unsupervised learning, given a contrastive encoder $f$, the primary objective is to make ensure that the embeddings of the positive pair ($x^+$, $x$), generated by the function $f$, are close to each other, while the embeddings of the negative pair ($x^-$, $x$), generated by the same function, are far away from each other.
Contrastive learning assumes access to similar data in the form of pairs ($x$, $x^+$) that come from a distribution $D_{sim}$ as well as $k$ i.i.d. negative samples $x_1^-, x_2^-, ..., x_k^-$ from a distribution $D_{neg}$ that are presumably unrelated to $x$. 
Learning is done over $\mathcal{F}$, a class of representation functions $f: \mathcal{X} \longrightarrow R^d$ where $f$ is the embedding function. 
The quality of the representation function $f$ (contrastive encoder) is evaluated by its performance on a multi-class classification task $\mathrm{T} \in \mathcal{T}$ using linear classification. 
A multi-class classifier for $\mathrm{T} \in \mathcal{T}$ is a function $g$  whose output coordinates are indexed by the classes $c$ in task $\mathrm{T} \in \mathcal{T}$.
For example, in SimCLR ~\cite{simCLR},
the encoder is denoted as $f$ and a linear classifier is used as the projector. So the whole framework of SimCLR can be expressed as $g(x) = w f(x)$. 
The loss considered in ~\cite{bound} is the logistic loss $l(v) = \textrm{log}_2 (1 + \sum_i \textrm{exp} (- v_i))$ for $v \in R^d$. Then the supervised loss of the downstream task classifier $g$ is 
\begin{align}
    L_{sup}(\mathrm{T}, g) = E [l( \{g(x)_c - g(x)_{c'} \}_{c \neq c'})]
\end{align}
where $c$ and $c'$ are different classes. 
For simplicity, we use $L_{sup}(f)$ to denote the downstream loss of the model with function $f$ which satisfies $L_{sup}(\mathrm{T}, g)$ and $g(x) = w f(x)$.

We outline the objective of contrastive learning: $k$ denotes number of negative samples used for training.
The unsupervised loss can be defined as 
\begin{align}
    L_{un}(f) = E[l(\{f(x)^T (f(x^+) - f(x^-)\}_{i=1}^k)]
\end{align}
After training, suppose $\hat{f}$ is the function which can minimizes the empirical unsupervised loss
and we denote its corresponding loss for supervised downstream task as $L_{sup}(\hat{f})$.
According to the theorim 4.1 in ~\cite{bound}, we have
\begin{align}\label{upper_bond}
    L_{sup}(\hat{f}) <= \alpha L_{un}(f) + \eta Gen_M + \epsilon
\end{align}

where $Gen$ is the generalization error which is defined as 
\begin{align}
Gen_M = O(R \sqrt{k} \frac{R_s (\mathcal{F})}{M} + (R^2 + log k) \sqrt{\frac{log \frac{1}{\epsilon}}{M}})
\end{align}
and $M$ is the sample size, $R_s (\mathcal{F})$ is the Rademacher average of $\mathcal{F}$ ~\cite{bound}, $\mathcal{F}$ is the function space defined by $f$, and $R$ is a constant which satisfies $||f(x)|| <= R$ for any $x$. 
This shows that the supervised task loss $L_{sup}(\hat{f})$ is bounded by the unsupervised loss, $L_{un}(f)$.

\noindent{\bf C - Skip-connection Based Model.}
Suppose we use neural network with skip connection ($f_2$) to learn the contrastive embedding, according to Eq.~\eqref{res_formula}, we have 
\begin{align}
L_{un}(f_2) &= E[l(\{f_2(x)^T (f_2(x^+) - f_2(x^-)\}_{i=1}^k)] \\
& = E[l(\{(f_I(x) + f_1(x))^T (f_I(x^+) + f_1(x^+) - f_I(x^-) - f_1(x^-)\}_{i=1}^k)] 
\end{align}
where $l$ is the logistic loss.
Suppose the learned $f_1(x)$ is a trivial identity matrix $\mathrm{I}$. As $x$ is closer to $x^+$ than to $x^-$, $f_I(x)^T (f_I(x^+) - f_I(x^-)) = f_I(x)^T f_I(x^+) - f_I(x)^T f_I(x^-) >= 0$ holds.
Accordingly, 
\begin{align*}
L_{un}(f_2) = L_{un}(f_I + f_I) &= E[l(4 \{f_I(x)^T (f_I(x^+) - f_I(x^-)\}_{i=1}^k)] \\
 & \leq E[l(\{f_I(x)^T (f_I(x^+) - f_I(x^-)\}_{i=1}^k)] = L_{un}(f_I) 
\end{align*} holds
because $l$ is monotonically decreasing. This means the upper bound of skip connection contrastive learning loss $L_{un}(f_2)$ is smaller than $L_{un}(f_I)$ which is the contrastive learning error of the original embedding.
If \myModel{} learns $f_1$ which is not an identity matrix through contrastive learning process, $L_{un}(f_2)$ is trivially less than $L_{un}(f_I + f_I)$, which induces that $L_{un}(f_2) \leq L_{un}(f_I)$ always holds. Therefore, the upper bound for using skip connections for contrastive learning should be lower.


We have observed that the proposed \myModel\ exhibits several similar properties to Adapter ~\cite{houlsby2019parameterefficient} in that both employ skip connection as their fundamental components as shown in Figure ~\ref{skip_connection}(d). However, Adapter is embedded within each layer of Transformer ~\cite{transformer}, while \myModel\ is positioned outside the original model $h()$. Although Adapter has been widely used in many Transformer ~\cite{transformer} based models ~\cite{bert,liu2019roberta}, no theoretical proof has been given thus far. In this work, we present the first theoretical proof demonstrating why skip connection-based refinement does not degrade downstream tasks.

\section{Experiment}\label{experiment}
Throughout the experiments, we want to show the effectiveness of \myModel{} through downstream task metric improvement and its wide applicability to various pre-trained embeddings over different modalities including shallow knowledge graph embedding, deep graph neural network embedding, image embedding, and text embeddings. The datasets and benchmark methods used in the study are initially described, followed by the presentation of experimental results. 

\subsection{Experimental Setting}
The study utilizes five datasets, as outlined below:
\begin{itemize}
  \item The \textbf{movieQA} is a movie knowledge graph derived from the WikiMovies Dataset. It includes over 40,000 triples that provide information about movies.
  \item The \textbf{STL10} is an image dataset for image classification task. It has 10 classes and has 500 96x96 training images along with 800 test image per class.
  \item The \textbf{CIFAR10} is an image dataset for image classification tasks. It has 10 classes and has 6,000 32x32 color images per class. The dataset is split into 50000 training images and 10000 test images.
  \item The \textbf{Cora} is a graph dataset for node classification. It consists of 2,708 scientific publications as nodes with seven classes and 5,429 citations as edges.
  \item The \textbf{Pubmed} is a graph dataset for node classification. It consists of 19,717 scientific publication in Pubmed as nodes with three classes and 44,338 citations as edges.
\end{itemize}

The following methods are employed to learn the input embedding for contrastive learning:

\begin{itemize}
    \item FedE ~\cite{fedE} is a Federated Knowledge Graph embedding framework that focuses on learning knowledge graph embeddings by aggregating locally-computed updates. For the local Knowledge Graph embedding, we employed TransE~\cite{transE}. This framework includes a client for each knowledge graph and a server for coordinating embedding aggregation.
    \item SimCLR ~\cite{simCLR} is a simple framework for contrastive learning of image representations. It first learns generic representations of images on an unlabeled dataset and then can be fine-tuned with a small number of labeled images to achieve good performance for a given classification task.
    \item GraphSAGE ~\cite{graphsage} is a general, inductive graph neural network (GNN) that leverages node feature information (e.g., text attributes). It samples and aggregates a nodes neighborhood's features to generate node embeddings.
    \item SimCSE ~\cite{gao2021simcse} is a self-supervised text embedding that refines any pre-training transformer-based language models. Its main idea is to apply contrastive learning by treating two text embeddings obtained from the same input text with different dropout as positive pairs.
\end{itemize}

\begin{figure}
    \centering
    \includegraphics[width=0.35\textwidth]{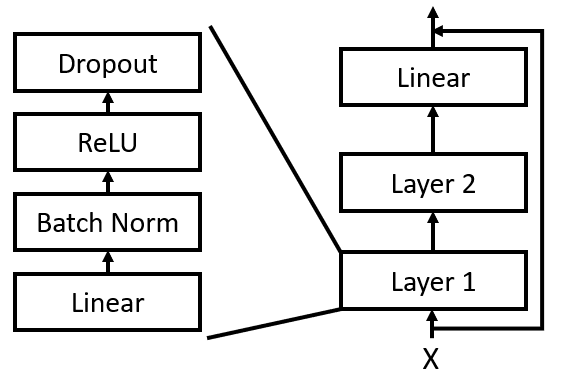}
    \caption{The SimSkip Encoder. Layer 1 and Layer 2 have the same architecture.}. 
    \label{architecture}
\end{figure}

Throughout the experiment, we adhere to the original baseline experiment settings when running baselines. The embedding dimensions are 128 for FedE, 128 for SIMCLR, 128 for GraphSage, and 768 for SIMCSE. As for our proposed \myModel, we explore various hyper-parameters over learning rate of 0.001, 0.0003, 000003, and 0.00001, and report its optimal performance. The encoder architecture of \myModel{} is shown in Figure ~\ref{architecture}. Layer 1 and layer 2 have the same structure, which contains a linear layer, a batch norm layer, a ReLU and a dropout layer. The projector is a two-layer feed forward network.  When the dimension of the original embedding is $d$, the number of parameters for the encoder is $d \times d/2$ for layer 1, $d/2 \times d$ for layer 2, and $d \times d$ for the linear layer. The number of parameters for the project is $d \times d$ for both layer 1 and layer 2. For data augmentation, the masking
augmentation randomly masked 20\% of the vector, and Gaussian noise augmentation
added noise sampled from a Gaussian distribution with mean 0 and variance 0.13.

\subsection{\myModel{} for Federated Knowledge Graph Embedding}

This section evaluates \myModel{}'s performance of refining the embedding learned by federated learning. To have the federated knowledge graph learning setting, two knowledge graphs are randomly sampled from the movieQA knowledge graph. FedE ~\cite{fedE} is used to learn the entity embeddings in a federated manner.


\begin{table}[h]
	\centering
	\caption{Accuracy on different downstream tasks for movieQA knowledge graph}
	\begin{tabular}{|c|c|c|c|c|c|c|}
	\hline
	        & KNN & \makecell{ Genre \\ Classification }  & \makecell{ Movie \\ Recommendation } \\ \hline
	FedE   & 58.5  & 84.8 &  7.47  \\ \hline
	\myModel + mask  & 62.8 & \textbf{86.1}   & \textbf{7.67} \\ \hline
        \myModel + gaussian  & \textbf{63.3} & 86.0   & 7.01 \\ \hline
	\end{tabular}
        \label{fede}
\end{table}

In this experiment, we use three different downstream tasks - k-nearest neighbor same genre prediction (KNN), genre classification and movie recommendation. KNN is that, given a query movie, take 10-nearest movies to the query movie in the embedding space and count how many movies in those 10 movies have the same genre as the query movie. Genre classification downstream task predicts the genre of a movie according to its embedding and a 3 layered MLP was employed as a downstream classfier. Movie recommendation task recommends new movies to users according to the user's watching history. The user's watching history data is from Netflix dataset \footnote{https://www.kaggle.com/code/laowingkin/netflix-movie-recommendation/data}. Given a user's watching history $V_1, ..., V_N$ in a chronological order, we treat $V_1, .., V_{N-10}$ as the training data, and predict 10 movies the user is most likely to watch. Then, we calculate how many movies in the 10 predicted movies belong to $V_{N-9}, ..., V_N$. 

The results of different methods are shown in Table ~\ref{fede}. \myModel\ improved accuracy of all three downstream tasks. For KNN task, the improvement is about 4\%. For genre classification and movie recommendation, the average improvement is 2\% and 0.2\%, respectively. 

\subsection{\myModel{} for Image Embedding}

In this section, we test \myModel{}'s performance of refining the self-supervised image embedding. We first use SimCLR ~\cite{simCLR} to learn the embedding, then we further refine the embedding with \myModel{}. STL10 and CIFAR10 datasets were used for evaluation. The downstream task is the image classification task and a 3-layer MLP was employed as a downstream classifier. Table ~\ref{contrastive_task} presents the downstream image classification accuracy and shows that \myModel{} refines the embedding space and improves the downstream task accuracy about 1\% in average.

\begin{table}[h]
	\centering
	\caption{Image classification accuracy on STL10 and CIFAR10}
	\begin{tabular}{|c|c|c|c|c|c|c|}
	\hline
	   Image Classification     &  STL10 &  CIFAR10  \\ \hline
	SimCLR   & 76.09  & 66.88   \\ \hline
	\myModel + mask  & 75.84  &  65.93 \\ \hline
        \myModel + gaussian  & \textbf{77.73}   &  \textbf{67.02}  \\ \hline
	\end{tabular}
        \label{contrastive_task}
\end{table}

\subsection{\myModel{} for Node embedding learned by Supervised Learning}

In this section, we test \myModel{}'s performance of refining the embedding learned by supervised learning. We use GraphSAGE ~\cite{graphsage} as the supervised embedding learner. Core and Pubmed were used for evaluation which GraphSAGE used for its evaluation. In the experiment, we first train GraphSAGE in supervised setting and treat the output of the second to last layer as the node embedding, then we apply \myModel{}. The downstream task is the node classification task and the same classification head of GraphSAGE was used as a downstream node classifier.

Table~\ref{contrastive_task_sage} shows that node classification accuracy on Cora and Pubmed datasets. Because we originally thought that the embedding trained by supervised learning should fit the downstream task best, further refining it with \myModel{} won't improve the downstream task performance. However, the experiment results show that \myModel{} further improved the downstream task performance even with the embedding learned by a supervised task.

\begin{table}[ht]
	\centering
	\caption{Node classification accuray on Cora and Pubmed}
	\begin{tabular}{|c|c|c|c|c|c|c|}
	\hline
	        & Cora & Pubmed  \\ \hline
	GraphSAGE   &  82.60 & 81.6   \\ \hline
	\myModel + mask  & 82.60  &  81.8 \\ \hline
        \myModel + gaussian  & \textbf{82.90}   &  \textbf{82.9}  \\ \hline
	\end{tabular}
        \label{contrastive_task_sage}
\end{table}

\subsection{\myModel{} for Transformer-based Text Embedding}

In this section, we test \myModel{}'s performance of refining the self-supervised transformer-based text embedding. Specifically, we further refine the pre-trained SimCSE ~\cite{gao2021simcse} embedding with \myModel{} and apply \myModel{} text embedding to various NLP downstream tasks including CR~\cite{CR}, MPQA ~\cite{MPQA}, MR ~\cite{MR}, MRPC ~\cite{MRPC}, SST-2 ~\cite{SST-2}, SUBJ ~\cite{SUBJ}, and TREC ~\cite{TREC}. These tasks are also used in ~\cite{gao2021simcse}. We use accuracy as the metric, which means that a higher value indicates better performance.
The results are presented in Figure ~\ref{simCSE_res}. Our findings suggest that stacking multiple embedding enhancing techniques (see SimCSE + \myModel{}) keeps improving the downstream task performance.


\begin{table*}
	\centering
	\caption{NLP task accuracy for self-supervised text embedding}
	\begin{tabular}{|c|c|c|c|c|c|c|c|c|c|c|}
	\hline
	Model         & CR & MPQA & MR  & MRPC  & SST2 & SUBJ & TREC \\ \hline
    SimCSE & 85.99 & 88.5  & 80.54 & 73.65 & \cellcolor{lightgray} 86.47 & 94.8  & \cellcolor{lightgray} 82.19 \\  \hline
    SimCSE + \myModel\ & \cellcolor{lightgray}  86.36 & 88.27 & \cellcolor{lightgray}  80.82 & 74.71 & 85.55 & 94.93 & 80.8  \\  \hline
    SimCSE + \myModel\ (mask+noise) & 85.44 & \cellcolor{lightgray} 88.56 & 78.25 & \cellcolor{lightgray} 75.01 & 82.96 & \cellcolor{lightgray} 95.28 & 80.1  \\  \hline
	
	\end{tabular}
	\label{simCSE_res}
\end{table*}

\subsection{Ablation Study}

In section, we assess the effect of the skip connection in \myModel{}. For comparison, we implemented $\myModel{}^{-}$ which is obtained by removing the skip connection from \myModel{}. For original embedding, we used embedding learned by SimCLR, STL10 and CIFAR10 as datasets, image classification as the downstream task. 
Table~\ref{contrastive_task_ablation} shows that the accuracy of the downstream task with $\myModel{}^{-}$ is lower than \myModel{} and even lower than that with the original SimCLR embedding (see Table~\ref{contrastive_task}). This aligns with our findings in subsection 3.1 which states that a wider and deeper network does not necessarily lead to a better approximation of the optimal function (see Figure~\ref{skip_connection} (b)). When the skip connection is omitted, the initial embedding obtained from the contrastive encoder becomes randomly dispersed across the entire embedding space. Consequently, subsequent updates have limited impact. However, with the inclusion of a skip connection, we ensure that the initial embedding from the contrastive encoder retains original useful information, facilitating the effectiveness of the subsequent updating process.

\begin{table}[h]
	\centering
	\caption{Ablation study of \myModel{} on image classification downstream task (unit: accuracy)}
	\begin{tabular}{|c|c|c|c|c|c|c|}
	\hline
	            &  STL10 &  CIFAR10  \\ \hline
	
	\myModel{} + mask  & 75.84  &  65.93 \\ \hline
        \myModel{} + gaussian  & \textbf{77.73}   &  \textbf{67.02}  \\ \hline
        $\myModel^{-}$  + mask & 47.1 & 56.5    \\ \hline
        $\myModel^{-}$ + gaussian & 56.6 & 66.7    \\ \hline
	\end{tabular}
        \label{contrastive_task_ablation}
\end{table}

\section{Related Work}\label{related_work}

\subsection{Representation Learning}

The goal of representation learning is to learn low dimension vectors of the input data so that similar data will be close to each other in the embedding space, while dissimilar data will be far from each other. It has been applied in many applications, such as dialogue system ~\cite{liu2023conversational,liu2023knowledge}, fact checking ~\cite{inspector,liu2022knowledge,liu2023knowledge,kompare}, language model ~\cite{lyu2022study,lyu2022multimodal,liu2024logic} and question answering ~\cite{newlook} and so on. 
For example, many language models use presentation learning to encode the input text information ~\cite{jin2022symlm,jin2023understand}, while many graph learning tasks utilizing graph neural network ~\cite{zhuang2022defending,zhuang2022does} to learn node representations. Other representation learning methods like TransE \cite{transE}, RESCAL ~\cite{RESCAL} and DistMult ~\cite{DistMult} embed entities in the konwledge graph as points in the low dimensional Euclidean and model relations as linear or bilinear transformation in the space.  Other methods like RotatE ~\cite{rotatE} and ComplEx ~\cite{complEx} represent entities as points in the complex space and relations as rotation or bilinear transfromation. Other methods like BoxE ~\cite{boxe} and KG2E ~\cite{gaussian_embedding} use geometry box or Gaussian distribution to represent an entity.

\subsection{Contrastive Learning}

Contrastive Learning focuses on minimizing the distance between the target embedding (anchor) vector and the matching (positive) embedding vector ~\cite{luo-etal-2022-kge,khosla2020supervised,you2020graph,wang-etal-2022-simkgc}, while maximizing the distance between the anchor vector and the non-matching (negative) embedding vectors. Recent work on contrastive learning have shown that discriminative or contrastive approaches can (i) produce transferable embeddings for visual objects through the use of data augmentation ~\cite{simCLR}, and (ii) learn joint visual and language embedding space that can be used to perform zero-shot detection ~\cite{zero}. Given the sparseness and long-tailed property of scene graph datasets, application (i) of contrastive approach can help the model learn better visual appearance embeddings of (subject, object) pairs under limited resource settings. Moreover, in application (ii), contrastive learning gives a clearer separation of the visual embeddings and language embeddings compared to the traditional black-box neural fusion approaches ~\cite{resnet}, giving us more control over both the symbolic triples input and the final output embedding spaces.

One thing we should notice is that the direct comparison of \myModel\ to other contrastive
techniques is not the primary focus. The main claim of \myModel\ is its ability to
further enhance the quality of embeddings learned by other contrastive methods
through a skip-connection based encoder-projector architecture with contrastive
learning in terms of downstream task performance. Accordingly, \myModel\ serves as
a facilitator of other techniques rather than a direct competitor. Additionally, since the embedding enhancing capability
of \myModel\ originates from the architecture rather than a specific contrastive
learning training technique, \myModel\ benefits from integrations with other
state-of-the-art contrastive learning techniques in various dimensions such as
loss function and data augmentation.



\section{Conclusion}\label{conclusion}
In this paper, we propose a skip connection based contrastive learning framework (\myModel{}) which refine the input embedding space. We theoretically prove that the downstream task error upper bounds with using \myModel{} embedding as its input will not be larger than that with the original embedding. The experiment results show the effectiveness of the proposed method. For future work, we intend to explore diverse data augmentation methods in embedding space and continue reducing the error bound in theoretical analysis. Besides, we plan to analyze how \myModel\ and related architectures can address the issues raised in Figure ~\ref{problem_of_existing_method}.


\bibliographystyle{splncs04}
\bibliography{reference}

\end{document}